\def\year{2019}\relax
\documentclass[letterpaper]{article} %DO NOT CHANGE THIS
\usepackage{aaai19}  %Required
\usepackage{times}  %Required
\usepackage{helvet}  %Required
\usepackage{courier}  %Required
\usepackage{url}  %Required
\usepackage{graphicx}  %Required
\usepackage{amsmath}
\usepackage{amssymb}
\usepackage{array}
\usepackage{multirow}
\usepackage{multicol}
\usepackage{bm}

\begin{document}
\title{Lattice CNNs for Matching Based Chinese Question Answering}
\author{Yuxuan Lai\textsuperscript{1}, Yansong Feng\textsuperscript{1,*}, Xiaohan Yu\textsuperscript{1} \\
      {\bf \Large Zheng Wang\textsuperscript{2}, Kun Xu\textsuperscript{3}, Dongyan Zhao\textsuperscript{1}} \\
\textsuperscript{1}Institute of Computer Science and Technology, Peking University, China\\
\textsuperscript{2}School of Computing and Communications, Lancaster University, UK ~~
\textsuperscript{3}Tencent AI Lab\\
\textsuperscript{1}\{erutan, fengyansong, yuxiaohan, zhaodongyan\}@pku.edu.cn\\
\textsuperscript{2}z.wang@lancaster.ac.uk 
\textsuperscript{3}syxu828@gmail.com \\
}
\maketitle

% Yuxuan Lai (Peking University) <erutan@pku.edu.cn>
% Yansong Feng (Peking University) <fengyansong@pku.edu.cn>
% Xiaohan Yu (Peking University) <1500012907@pku.edu.cn>
% Zheng Wang (Lancaster University) <z.wang@lancaster.ac.uk>
% Kun Xu (Tencent AI lab) <xukun@pku.edu.cn>
% Dongyan Zhao (Peking Univeristy) <zhaodongyan@pku.edu.cn>

\begin{abstract}
Short text matching often faces the challenges that there are great word mismatch and expression diversity between the two texts, 
which would be further aggravated in languages like Chinese where there is no natural space to segment words explicitly.
In this paper,  we propose a novel lattice based CNN model  (LCNs)  to utilize multi-granularity information inherent in the word lattice while maintaining 
strong ability to deal with the introduced noisy information for matching based question answering in Chinese.
We conduct extensive experiments on both document based question answering and knowledge based question answering tasks, and experimental 
results 
show that the LCNs models can significantly outperform the state-of-the-art matching models and strong baselines by taking advantages 
of better ability to distill rich but discriminative information from the word lattice input. \footnote{Code and appendix can be found in https://github.com/Erutan-pku/LCN-for-Chinese-QA .}

\end{abstract}

\section{Introduction}
\begin{figure}[t]
    \centering
    \includegraphics[width=0.48\textwidth]{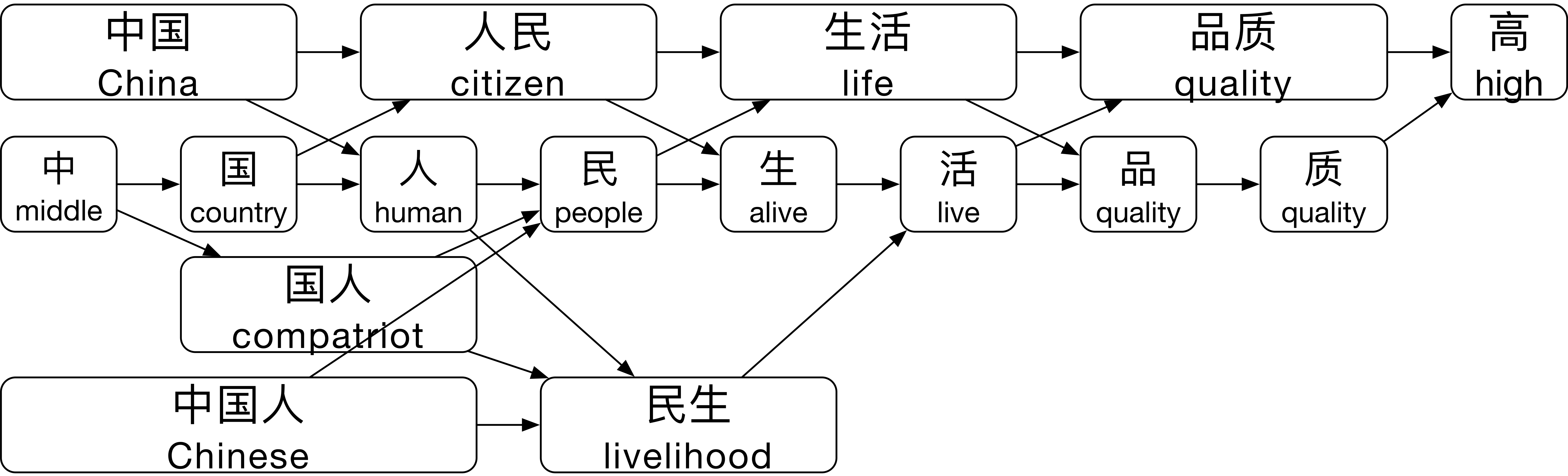}
\caption{ A word lattice for the phrase ``Chinese people have high quality of life."
        \label{fig:lattice}}
\end{figure}

Short text matching plays a critical role in many natural language processing tasks, such as question answering, information retrieval, and so on. 
However, matching text sequences for Chinese or similar languages often suffers from word segmentation, where there are often no perfect Chinese word segmentation tools that suit every scenario. Text matching usually requires to capture the relatedness between two sequences in multiple granularities. 
For example, in Figure~\ref{fig:lattice}, the example phrase is generally tokenized as ``China – citizen – life – quality – high'', but when we plan to match it with ``Chinese – live – well'', it would be more helpful to have the example segmented into ``Chinese – livelihood – live'' than its common segmentation.
\footnote{For clarity, \emph{``Italic''} are examples organised in Chinese Pinyin followed by its translation in English, and ``–'' represents a separator between Chinese words.}

Existing efforts use neural network models to improve the matching based on the fact that distributed representations can generalize discrete word features in traditional bag-of-words methods.
And there are also works  fusing word level and character level information, which, to some extent, could relieve the mismatch between different  segmentations, but these solutions 
still suffer from the original word sequential structures. They usually depend on an existing word tokenization, which has to make segmentation choices at one time, e.g., \emph{``ZhongGuo''}(China) and \emph{``ZhongGuoRen''}(Chinese) when processing \emph{``ZhongGuoRenMin''}(Chinese people).
And the blending just conducts at one position in their frameworks.

Specific tasks such as question answering (QA) could pose further challenges for short text matching. 
In document based question answering (DBQA), the matching degree is expected to reflect how likely a sentence can answer a given question, where questions and candidate answer sentences usually come from different sources, and may exhibit significantly different styles or syntactic structures, e.g. queries in web search and sentences in web pages.
This could further aggravate the mismatch problems.
In knowledge based question answering (KBQA), one of the key tasks is to match relational expressions in questions with knowledge base (KB) predicate phrases\footnote{There are usually not enough training data to build a relation extractor for each predicate in a KB, thus the task of KB predicate identification is often formulated as a matching task, which is to select predicates that match the given questions from the candidates.}, such as \emph{``ZhuCeDi''}(place of incorporation). Here the diversity between the two kinds of expressions is even more significant, where there may be dozens of different verbal expressions in natural language questions corresponding to only one KB predicate phrase. 
Those expression problems make KBQA a further tough task.
Previous works \cite{sa_qa2,Yih2015Semantic} adopt letter-trigrams for the diverse expressions, which is similar to character level of Chinese. And the lattices are combinations of words and characters, so with lattices, we can utilize words information at the same time.

Recent advances have put efforts in modeling multi-granularity information for matching.
\cite{bidaf,sa_match1} blend words and characters to a simple sequence (in word level), and \cite{chen2018mix} utilize multiple convoluational kernel sizes to capture different n-grams.
But most characters in Chinese can be seen as words on their own, 
so combining characters with corresponding words directly may lose the meanings that those characters can express alone.
Because of the sequential inputs, they will either lose word level information when conducting on character sequences or have to make segmentation choices.

In this paper, we propose a multi-granularity method for short text matching in Chinese question answering which utilizes lattice based CNNs to extract sentence level features over word lattice.
Specifically, instead of relying on character or word level sequences, LCNs take word lattices as input, 
where every possible word and character will be treated equally and have their own context
so that they can interact at every layer. 
For each word in each layer, LCNs can capture different context words in different granularity via pooling methods.
To the best of our knowledge, we are the first to introduce word lattice into the text matching tasks. 
Because of the similar IO structures to original CNNs and the high efficiency,  LCNs can be easily 
adapted to more scenarios where flexible sentence representation modeling is required.

We evaluate our LCNs models on two question answering tasks,  document based question answering and knowledge based question answering, both in Chinese. 
Experimental results show that LCNs significantly outperform the state-of-the-art matching methods and other competitive CNNs baselines in both scenarios. We also find that LCNs can better capture the multi-granularity information
from plain sentences, and, meanwhile,  maintain better de-noising capability than vanilla graphic convolutional neural networks thanks to its dynamic convolutional kernels and gated pooling mechanism.

\section{Lattice CNNs}
Our Lattice CNNs framework is built upon the siamese architecture \cite{siamese}, one of the most successful frameworks in text matching, which takes the word lattice format of a pair of sentences as input, and outputs the matching score. 

\subsection{Siamese Architecture}
The siamese architecture and its variant have been widely adopted in sentence matching \cite{sa_match2,sa_match1} and matching based question answering \cite{sa_qa1,sa_qa2,sa_qa3}, that has a symmetrical component to extract high level features from different input channels, which share parameters and map inputs to the same vector space. Then, the sentence representations are merged and compared to output the similarities.

\begin{figure*}[t]
    \centering
    \includegraphics[width=0.85\textwidth]{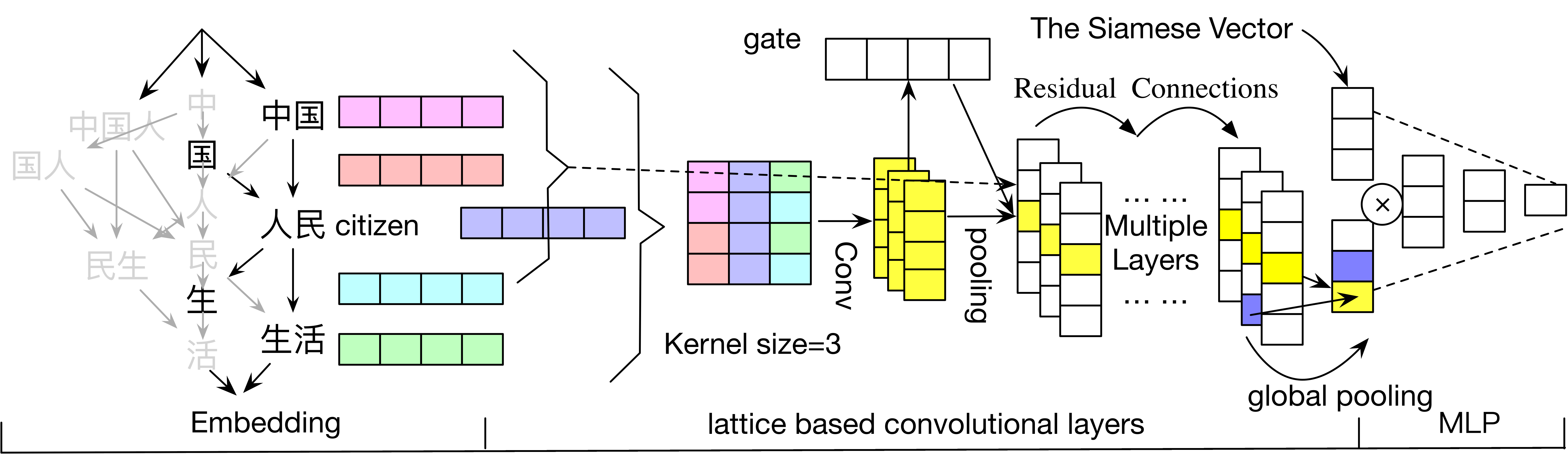}
\caption{ An illustration of our LCN-gated, when ``people'' is being considered as the center of convolutional spans.
        \label{fig:architecture}}
\end{figure*}

For our models, we use multi-layer CNNs for sentence representation. Residual connections \cite{residual} are used between convolutional layers to enrich features and make it easier to train. 
Then, max-pooling summarizes the global features to get the sentence level representations, which are merged via element-wise multiplication.
The matching score is produced by a multi-layer perceptron (MLP) with one hidden layer based on the merged vector. The fusing and matching procedure is formulated as follows:
\begin{equation}
s = \sigma(\bm{W}_2~\texttt{ReLU}(\bm{W}_1(\bm{f}_{qu}\odot \bm{f}_{can})+\bm{b}_1^T)+\bm{b}_2^T)
\end{equation}
where $\bm{f}_{qu}$ and $\bm{f}_{can}$ are feature vectors of question and candidate (sentence or predicate) separately encoded by CNNs, $\sigma$ is the sigmoid function, $\bm{W_2, W_1, b_1^T, b_2^T}$ are parameters, and $\odot$ is element-wise multiplication.
The training objective is to minimize the binary cross-entropy loss, defined as:
\begin{equation}
L=-\sum_{i=1}^{N}{[y_i log(s_i)+(1-y_i) log(1-s_i)]}
\end{equation}
where $y_i$ is the \{0,1\} label for the $i_{th}$ training pair.

Note that the CNNs in the sentence representation component can be either original CNNs with sequence input or lattice based CNNs with lattice input. 
Intuitively, in an original CNN layer, several kernels scan every n-gram in a sequence and result in one feature vector, which can be seen as the representation for the center word and will be fed into the following layers. 
However, each word may have different context words in different granularities in a lattice and 
may be treated as the center in various kernel spans with same length. 
Therefore, different from the original CNNs, 
there could be several feature vectors produced for a given word, 
which is the key challenge to apply the standard CNNs directly to a lattice input. 

For the example shown in Figure~\ref{fig:architecture}, the word ``citizen'' is the center word of four text spans with length 3: ``China - citizen - life'', ``China - citizen - alive'', ``country - citizen - life'', ``country - citizen - alive'', so four feature vectors will be produced for width-3 convolutional kernels  for ``citizen''.

\subsection{Word Lattice}
As shown in Figure~\ref{fig:lattice}, a word lattice is a directed graph $G = \langle V,E \rangle $, where $V$ represents a node set and $E$ represents a edge set. For a sentence in Chinese, which is a sequence of 
Chinese characters $S=c_{1:n}$, all of its possible substrings that can be considered 
% =c_1,c_2,...,c_n
as words  are treated as vertexes, i.e. $V = \{c_{i:j} | c_{i:j} ~\texttt{is word}\} $. Then, all neighbor words are connected by directed edges according to their positions in the original sentence, i.e. $E = \{ e(c_{i:j}, c_{j:k}) | \forall i, j, k ~~ s.t. ~ c_{i:j}, c_{j:k} \in V \}$. 

Here, one of the key issues is how we decide a sequence of characters can be considered as a word.
We approach this through an existing lookup vocabulary, %according to the word vector vocabulary, 
which contains frequent words in BaiduBaike\footnote{https://baike.baidu.com}.
Note that most Chinese characters can be considered as words on their own, thus are included in this vocabulary when they have been used as words on their own in this corpus.

However, doing so will inevitably introduce noisy words (e.g., ``middle'' in Figure~\ref{fig:lattice}) into word lattices, which will be smoothed by pooling procedures in our model. 
And the constructed graphs could be disconnected because of a few out-of-vocabulary characters. Thus, we append $\langle unk\rangle$ labels to replace those characters to connect the graph.

Obviously, word lattices are collections of characters and all possible words.
Therefore, it is not necessary to make explicit decisions regarding specific  
word segmentations, but just embed all possible information into the lattice and take them to the next CNN layers. 
The inherent graph structure of a word lattice allows all possible words represented explicitly, no matter the overlapping and nesting cases, and all of them can contribute directly to the sentence representations.

\subsection{Lattice based CNN Layer}
As we mentioned in previous section, we can not directly apply standard CNNs to take word lattice as input, 
since there could be multiple  feature vectors produced for a given word.
Inspired by previous lattice LSTM models\cite{latticeLSTMnmt,latticeLSTMner},  
here we propose a lattice based CNN layers to allow standard CNNs to work over word lattice input. 
Specifically, we utilize pooling mechanisms to merge the feature vectors produced by multiple CNN kernels over different context compositions. 

Formally, the output feature vector of a lattice CNN layer with kernel size $n$ at word $w$ in a word lattice $G = \langle V,E \rangle$ can be formulated as Eq~\ref{lattice_cnn} :

\begin{equation}
\begin{aligned} 
F_{w} = g \{ f(\bm{W}_c(\bm{v}_{\bm{w}_{1}}:...:\bm{v}_{\bm{w}_{n}}) + \bm{b}_c^{T}) | \\
\forall i , w_{i} \in V, (w_{i}, w_{i+1}) \in E, w_{\left\lceil\frac{n+1}{2}\right \rceil} =  w \} \\
\end{aligned}
\label{lattice_cnn}
\end{equation}
where $f$ is the activation function, $\bm{v}_{\bm{w}_{i}}$ is the input vector corresponding to word $w_{i}$ in this layer, ($\bm{v}_{\bm{w}_{1}}:...:\bm{v}_{\bm{w}_{n}})$ means the concatenation of these vectors, and $\bm{W}_c, \bm{b}_c$ are parameters with size $[m', n \times m]$, and $[m']$, respectively. $m$ is the input dim and $m'$ is the output dim. $g$ is one of the following pooling functions: max-pooling, ave-pooling, or gated-pooling, which execute the element-wise maximum, element-wise average, and the gated operation, respectively. The gated operation can be formulated as:
\begin{eqnarray}
\alpha_1, ...,\alpha_t = \texttt{softmax}\{\bm{v}_g^T \bm{v}_1+b_g, ..., \bm{v}_g^T \bm{v}_t+b_g\} \\
\texttt{gated-pooling}\{\bm{v}_1, ..., \bm{v}_t\}=\sum_{i=1}^{n}{\alpha_i \times \bm{v}_i} 
\label{gated_pooling}
\end{eqnarray}
where $\bm{v}_g, b_g$ are parameters, and $\alpha_i$ are gated weights normalized by a softmax function. Intuitively, the gates represent the importance of the n-gram contexts, and the weighted sum can control the transmission of noisy context words.
We perform padding when necessary.

For example, in Figure~\ref{fig:architecture}, when we consider ``citizen'' as the center word, and the kernel size is 3, there will be five words and four context compositions involved, as mentioned in the previous section, each marked in different colors. Then, 3 kernels scan on all compositions and produce four 3-dim feature vectors. The gated weights are computed based on those vectors via a dense layer, which can reflect the importance of each context compositions. The output vector of the center word is their weighted sum, where noisy contexts are expected to have lower weights to be smoothed. This pooling over different contexts allows LCNs to work over word lattice input.

Word lattice can be seen as directed graphs and modeled by Directed Graph Convolutional networks (DGCs) \cite{dgc1}, which use poolings on neighboring vertexes that ignore the semantic structure of n-grams. But to some situations, their formulations can be very similar to ours (See \textbf{Appendix}\footnote{https://github.com/Erutan-pku/LCN-for-Chinese-QA/blob/master/paper\_appendix.pdf } for derivation). For example, if we set the kernel size in LCNs to 3, use linear activations and suppose the pooling mode is average in both LCNs and DGCs, at each word in each layer, the DGCs compute the average of the first order neighbors together with the center word, while the LCNs compute the average of the pre and post words separately and add them to the center word. Empirical results are exhibited in \textbf{Experiments} section.

Finally, given a sentence that has been constructed into a word-lattice form, for each node in the lattice, an LCN layer will produce one feature vector similar to original CNNs, which makes it easier to stack multiple LCN layers to obtain more abstract feature representations.

\section{Experiments}

Our experiments are designed to answer: (1) whether multi-granularity information in word lattice helps in matching based QA tasks, (2) whether LCNs capture the multi-granularity information through lattice well, and (3) how to balance the noisy and informative words introduced by word lattice.

\subsection{Datasets}
We conduct experiments on two Chinese question answering datasets from NLPCC-2016 evaluation task \cite{overview2016nlpcc}.

\textbf{DBQA} is a document based question answering dataset. 
There are 8.8k questions with 182k question-sentence pairs for training and 6k questions with 123k question-sentence pairs in the test set. In average, each question has 20.6 candidate sentences and 1.04 golden answers. The average length for questions is 15.9 characters, and each candidate sentence has averagely 38.4 characters. 
Both questions and  sentences are natural language sentences, possibly sharing more similar word choices and expressions compared to the KBQA case. But the candidate sentences are extracted from web pages, and are often much longer than the questions,  with many irrelevant clauses.

\textbf{KBRE} is a knowledge based relation extraction dataset. 
We follow the same preprocess as \cite{nlpcckbqa2017top1} to clean the dataset\footnote{About 3\% of the questions in the original dataset are removed for they can not link to correct entities/relations due to label errors.}  and replace entity mentions in questions to a special token. 
There are 14.3k questions with 273k question-predicate pairs in the training set and 9.4k questions with 156k question-predicate pairs for testing. Each question contains  only one golden predicate. 
Each question averagely has 18.1 candidate predicates and 8.1 characters in length, while a KB predicate is  only 3.4 characters long on average. 
Note that a KB predicate is usually a concise phrase, with quite different word choices
compared to the natural language questions, which poses different challenges to solve.

The vocabulary we use to construct word lattices contains 156k words, including 9.1k single character words.
In average, each DBQA question contains 22.3 tokens (words or characters) in its lattice, each DBQA candidate sentence has 55.8 tokens, each KBQA question has 10.7 tokens  and each KBQA predicate contains 5.1 tokens.

\subsection{Evaluation Metrics}
For both datasets, we follow the evaluation metrics used in the original evaluation tasks \cite{overview2016nlpcc}. For DBQA, P@1 (Precision@1), MAP (Mean Average Precision) and MRR (Mean Reciprocal Rank) are adopted. For KBRE, since only one golden candidate is labeled for each question, only P@1 and MRR are used. %We omit any further details for the sake of brevity.

\subsection{Implementation Details}
The word embeddings are trained on the Baidu Baike webpages with Google's word2vector\footnote{https://code.google.com/archive/p/word2vec/}, which are 300-dim and fine tuned during training. In DBQA, we also follow previous works \cite{nlpccdbqa2016top1,nlpccdbqa2017top1} to concatenate additional 1d-indicators with word vectors which denote whether the words are concurrent in both questions and candidate sentences. In each CNN layer, there are 256, 512, and 256 kernels with width 1,  2, and  3, respectively. 
The size of the hidden layer for MLP is 1024. All activation are ReLU, the dropout rate is 0.5, with a batch size of 64. We optimize with adadelta \cite{adadelta} with learning rate $=1.0$ and decay factor $=0.95$. We only tune the number of convolutional layers from [1, 2, 3] and fix other hyper-parameters.  We sample at most 10 negative sentences per question in DBQA and 5 in KBRE. We implement our models in Keras\footnote{https://keras.io} with Tensorflow\footnote{https://www.tensorflow.org} backend.

\subsection{Baselines}

Our first set of baselines uses original CNNs with character (\textbf{CNN-char}) or word inputs. For each sentence, two Chinese word segmenters are used to obtain three different word sequences: jieba (\textbf{CNN-jieba})\footnote{https://pypi.python.org/pypi/jieba/}, and Stanford Chinese word segmenter\footnote{https://nlp.stanford.edu/software/segmenter.shtml} in CTB (\textbf{CNN-CTB}) and PKU (\textbf{CNN-PKU}) mode. 

Our second set of baselines combines
different word segmentations. Specifically, we concatenate the sentence embeddings from different segment results, 
which gives  four different word+word models: \textbf{jieba+PKU}, \textbf{PKU+CTB}, \textbf{CTB+jieba}, and \textbf{PKU+CTB+jieba}. 

Inspired by previous works \cite{bidaf,sa_match1}, we also concatenate word and character embeddings at the input level. Specially, when the basic sequence is in word level, each word may be constructed by multiple characters through a pooling operation (\textbf{Word+Char}). 
Our pilot experiments show that average-pooling is the best for DBQA while max-pooling after a dense layer is the best for KBQA. 
When the basic sequence is in character level, we simply concatenate the character embedding with its corresponding word embedding (\textbf{Char+Word}), since each character belongs to
one word only. 
Again, when the basic sequence is in character level, 
we can also concatenate the character embedding with a pooled representation of all words 
that contain this character 
in the word lattice (\textbf{Char+Lattice}), where we use max pooling as suggested by our pilot experiments.

DGCs \cite{dgc1,dgc3} are strong baselines that perform CNNs over directed graphs
to produce high level representation for each vertex in the graph, which can be used to
build a sentence representation via certain pooling operation.  
We therefore choose to compare with \textbf{DGC-max} (with maximum pooling), \textbf{DGC-ave} (with average pooling), 
and \textbf{DGC-gated} (with gated pooling), 
where the gate value is computed using the concatenation of the vertex vector and the center vertex vector through a dense layer. % 

We also implement several state-of-the-art matching models using the open-source project MatchZoo \cite{matchzoo}, where we tune hyper-parameters using grid search, e.g., whether using word or character inputs.
\textbf{Arc1, Arc2, CDSSM} are traditional CNNs based matching models proposed by \cite{matchzooARC,matchzooCDSSM}. Arc1 and CDSSM compute the similarity via sentence representations and Arc2 uses the word pair similarities.
\textbf{MV-LSTM} \cite{matchzooMVLSTM} computes the matching score by examining the interaction between the representations from two sentences 
obtained by a shared BiLSTM encoder. 
\textbf{MatchPyramid(MP)} \cite{matchzooMP} utilizes 2D convolutions and pooling strategies over word pair similarity matrices
to compute the matching scores. 

We also compare with the state-of-the-art models in DBQA \cite{nlpccdbqa2016top1,nlpccdbqa2017top1}.

\subsection{Results}
\begin{table}[t]
\footnotesize
\centering
\begin{tabular}{l|lll|ll}
\multicolumn{1}{c}{\multirow{2}{*}{}} & \multicolumn{3}{c}{DBQA}                                                    & \multicolumn{2}{c}{KBRE}                          \\
\multicolumn{1}{c}{}      & \multicolumn{1}{c}{MAP} & \multicolumn{1}{c}{MRR} & \multicolumn{1}{c}{P@1} & \multicolumn{1}{c}{P@1} & \multicolumn{1}{c}{MRR} \\ \hline\hline
 \multicolumn{6}{c}{MatchZoo}                        \\ \hline
Arc1                      & .4006        & .4011        & 22.39\%        & 32.18\%        & .5144        \\
Arc2                      & .4780        & .4785        & 30.47\%        & 76.07\%        & .8518        \\
CDSSM                     & .5344        & .5349        & 36.45\%        & 68.90\%        & .7974        \\
MP                        & .7715        & .7723        & 65.61\%        & 86.21\%        & .9137        \\
MV-LSTM                   & \bf.8154     & \bf.8162     & \bf71.71\%     & \bf86.87\%     & \bf.9271     \\ \hline
\multicolumn{6}{c}{State-of-the-Art DBQA}                        \\ \hline
(Fu et al. 2016)          & .8586        & .8592        & 79.06\%        & ---            & ---          \\ 
\cite{nlpccdbqa2017top1}* & \bf.8763     & \bf.8768     & ---            & ---            & ---          \\ \hline
\multicolumn{6}{c}{Single Granularity CNNs}                        \\ \hline
CNN-jieba                 & .8281        & .8289        & 75.10\%        & 86.85\%        & .9152        \\ 
CNN-PKU                   & .8339        & .8343        & 76.00\%        & 89.87\%        & .9370        \\ 
CNN-CTB                   & .8341        & .8347        & 76.04\%        & 88.92\%        & .9302        \\
CNN-char                  & \bf.8803     & \bf.8809     & \bf82.09\%     & \bf93.06\%     & \bf.9570     \\ \hline 
\multicolumn{6}{c}{Word Combine CNNs}                        \\ \hline
jieba+PKU                 & .8486        & .8490        & 77.62\%        & 90.57\%        & .9417        \\ 
PKU+CTB                   & .8435        & .8440        & 77.09\%        & 90.48\%        & .9410        \\ 
CTB+jieba                 & \bf.8499     & \bf.8504     & \bf78.06\%     & 90.29\%        & .9399        \\ 
\scriptsize PKU+CTB+jieba & .8494        & .8498        & 78.04\%        & \bf91.16\%     & \bf.9450     \\ \hline 
\multicolumn{6}{c}{Word+Char CNNs}                        \\ \hline
Word+Char                 & .8566        & .8570        & 78.94\%        & 91.64\%        & .9489        \\
Char+Word                 & .8728        & .8735        & 80.76\%        & 92.78\%        & .9561        \\
Char+Lattice              & \bf.8810     & \bf.8815     & \bf81.97\%     & \bf93.12\%     & \bf.9582     \\ \hline
\multicolumn{6}{c}{DGCs}                        \\ \hline
DGC-ave                   & \bf.8868     & \bf.8873     & \bf83.02\%     & \bf93.49\%     & \bf.9602     \\
DGC-max                   & .8811        & .8818        & 82.01\%        & 92.79\%        & .9553        \\
DGC-gated                 & .8790        & .8795        & 81.69\%        & 92.88\%        & .9562        \\ \hline
\multicolumn{6}{c}{LCNs}                        \\ \hline
LCN-ave                   & .8864        & .8869        & 83.14\%        & \bf93.60\%     & \bf.9609     \\
LCN-max                   & .8870        & .8875        & 83.06\%        & 93.54\%        & .9604        \\
LCN-gated                 & \bf.8895     & \bf.8902     & \bf83.24\%     & 93.32\%        & .9592            

\end{tabular}
\caption{The performance of all models on the two datasets. The best results in each group are bolded. * is the best published DBQA result. 
        \label{overall_performance}}
\end{table}

Here, we mainly describe the main results on the DBQA dataset, while we find very similar trends on the KBRE dataset.
Table~\ref{overall_performance} summarizes the main results on the two datasets. We can see 
that the simple MatchZoo models perform the worst.
Although Arc1 and CDSSM are also constructed in the siamese architecture with CNN layers, they do not employ  
multiple kernel sizes and residual connections, and fail to capture the relatedness in a multi-granularity fashion. 

\cite{nlpccdbqa2016top1} is similar to our word level models (CNN-jieba/PKU/CTB), but outperforms  our models by around 3\%, 
 since it benefits from an extra interaction layer with fine tuned hyper-parameters. 
\cite{nlpccdbqa2017top1} further incorporates human designed features including POS-tag interaction and TF-IDF scores, achieving state-of-the-art 
performance in the literature of this DBQA dataset. 
However, both of them perform worse than our simple \textbf{CNN-char} model, which is a strong baseline because characters, that describe the text in a fine granularity, can relieve word mismatch problem to some extent. 
And our best LCNs model further outperforms \cite{nlpccdbqa2017top1} by .0134 in MRR.   

For single granularity CNNs,  \textbf{CNN-char} performs  better than all word level models, because they heavily suffer from word mismatching given one fixed 
word segmentation result. 
And the models that utilize different word segmentations can relieve this problem and gain better performance, which can be further improved by the combination of words and characters. % because the complementary granularity.

The DGCs and LCNs, being able to work on lattice input, outperform all previous models that have sequential inputs,
\footnote{The best LCN models can reduce the error rates (= 1 - P@1) over \textbf{Char+Lattice} by 7.04\% in DBQA and 6.98\% in KBRE.}
indicating that the word lattice is a more promising form than a single word sequence, and  should be better captured by taking  the inherent graph structure into account. 
Although they take the same input, LCNs still perform better than the best DGCs by a margin, showing the advantages of the CNN kernels over multiple n-grams in the lattice structures and the gated pooling strategy.

To fairly compare with previous KBQA works, we combine our LCN-ave settings with the entity linking results of the state-of-the-art KBQA model\cite{nlpcckbqa2017top1}. The P@1 for question answering of single LCN-ave is 86.31\%, which outperforms both the best single model (84.55\%) and the best ensembled model (85.40\%) in literature.

\subsection{Analysis and Discussions}

\subsubsection{Effectiveness of Multi-Granularity information}
As shown in Table~\ref{overall_performance}, the combined word level models  (e.g. \textbf{CTB+jieba} or \textbf{PKU+CTB}) perform better than 
any word level CNNs with single word segmentation result (e.g. \textbf{CNN-CTB} or \textbf{CNN-PKU}).  
The main reason is that there are often no perfect Chinese word segmenters and a single improper segmentation decision may
harm the matching performance, since that could  further make the word mismatching issue worse, 
while  the combination of different word segmentation results can somehow relieve this situation.

Furthermore, the models combining words and characters all perform better than \textbf{PKU+CTB+jieba}, 
because they could be complementary in different granularities. 
Specifically, \textbf{Word+Char} is still worse than \textbf{CNN-char}, because Chinese characters have rich meanings and compressing several characters to a single word vector will inevitably lose information. Furthermore, the combined sequence of \textbf{Word+Char} still exploits in a word level, 
which still suffers from the single segmentation decision. 
On the other side, the \textbf{Char+Word} model is also slightly worse than \textbf{CNN-char}. We think one reason is that the reduplicated word embeddings concatenated with each character vector confuse the CNNs, and perhaps lead to overfitting. 
But, we can still see that  \textbf{Char+Word} performs better than \textbf{Word+Char}, because the former exploits in a character level and the fine-granularity information actually helps to relieve word mismatch.
Note that \textbf{Char+Lattice}  outperforms \textbf{Char+Word}, and even slightly better than \textbf{CNN-char}.
This illustrates that multiple word segmentations are still helpful to further improve the character level strong baseline \textbf{CNN-char}, which may still benefit from word level information in a multi-granularity fashion.

In conclusion, the combination between different sequences and information of different granularities can help improve text matching, showing that it is necessary to consider the fashion which considers both characters and more possible words, which perhaps the word lattice can provide.

\subsubsection{Poolings in DGCs and LCNs}

For DGCs with different kinds of pooling operations, average pooling (\textbf{DGC-ave}) performs the best, which delivers similar performance with \textbf{LCN-ave}. %, which confirms the fact that they have similar formulations. 
While \textbf{DGC-max} performs a little worse, because it ignores the importance of different edges and the maximum operation is more sensitive to noise than the average operation.
The \textbf{DGC-gated} performs the worst. Compared with \textbf{LCN-gated} that learns the gate value adaptively from multiple n-gram context, it is harder for DGC to learn the importance of each edge via the node and the center node in the word lattice.
It is not surprising that \textbf{LCN-gated} performs much better than \textbf{GDC-gated}, indicating again that n-grams in word lattice play an important role in context modeling, while DGCs are designed for general directed graphs which may not be perfect to work with word lattice.

For LCNs with different pooling operations, \textbf{LCN-max} and \textbf{LCN-ave} lead to similar performances, and perform better on KBRE, while \textbf{LCN-gated} is better on DBQA. 
This may be due to the fact that sentences in DBQA are relatively longer with more irrelevant information
which require
% \textbf{LCN-gated} 
to filter noisy context, while on KBRE with much shorter predicate phrases, \textbf{LCN-gated} may slightly overfit due to its more complex model structure.
Overall, we can see that LCNs perform better than DGCs, thanks to the advantage of better capturing multiple n-grams context in word lattice.   

\subsubsection{How LCNs utilizes Multi-Granularity} 

\begin{figure}[]
    \centering
    \includegraphics[width=0.47\textwidth]{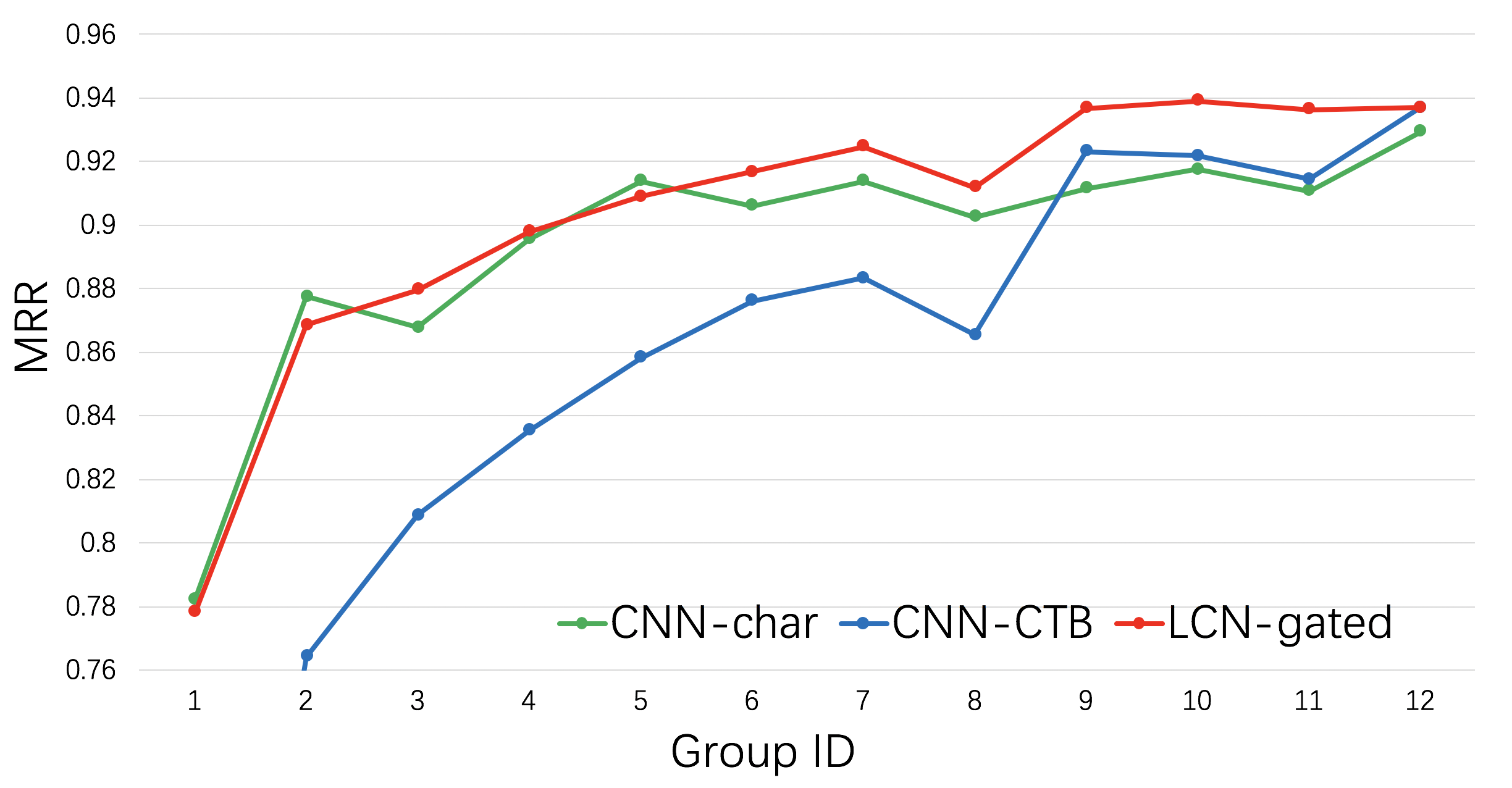}
    
\caption{ MRR score against granularities of overlaps between questions and answers, which is the average length of longest common substrings. About 2.3\% questions are ignored for they have no overlaps and the rests are separated in 12 groups orderly and equally. Group 1 has the least average overlap length while group 12 has the largest. 
        \label{fig:analysis}}
\end{figure}

To investigate how LCNs utilize multi-granularity more intuitively,
we analyze the MRR score against granularities of overlaps between questions and answers in DBQA dataset, which is shown in Figure~\ref{fig:analysis}. 
It is demonstrated that \textbf{CNN-char} performs better than \textbf{CNN-CTB} impressively in first few groups where most of the overlaps are single characters which will cause serious word mismatch. 
With the growing of the length of overlaps, \textbf{CNN-CTB} is catching up and finally overtakes \textbf{CNN-char} even though its overall performance is much lower. 
This results show that word information is complementary to characters to some extent.
The \textbf{LCN-gated} is approaching the \textbf{CNN-char} in first few groups, and outperforms both character and word level models in next groups, where word level information becomes more powerful.
This demonstrates that LCNs can effectively take advantages of different granularities, and the combination will not be harmful even when the matching clues present in extreme cases.

\textbf{How to Create Word Lattice}  ~~ 
In previous experiments, we construct word lattice via an existing lookup vocabulary, which will introduce some noisy words inevitably. 
Here we construct from various word segmentations with different strategies to investigate the balance between the noisy words and additional information introduced by word lattice. 
We only use the DBQA dataset because word lattices here are more complex, so the construction strategies have more influence. 
Pilot experiments show that word lattices constructed based on character sequence perform better, so the strategies in Table~\ref{lattcie_construct} are based on \textbf{CNN-char}.

From Table~\ref{lattcie_construct}, it is shown that all kinds of lattice are better than \textbf{CNN-char}, which also evidence the usage of word information. 
And among all LCN models, more complex lattice produces better performance in principle, which indicates that LCNs can handle the noisy words well and the influence of noisy words can not cancel the positive information brought by complex lattices.
It is also noticeable that \textbf{LCN-gated} is better than LCN-C+20 by a considerable margin, which shows that the words not in general tokenization (e.g. ``livelihood'' in Fig~\ref{fig:lattice}) are potentially useful.

\begin{table}[]
\centering
\begin{tabular}{l|cccc}
          & MRR   & P@1     & l.qu & l.can \\ \hline\hline
CNN-char  & .8809 & 82.09\% & 15.9 & 38.4  \\ \hline
LCN-C+2\& & .8851 & 82.41\% & 19.9 & 48.0  \\
LCN-C+2   & .8874 & 82.89\% & 20.4 & 49.5  \\
% LCN-C+20f & .8868 & 82.82\% & 20.8 & 49.9  \\
LCN-C+20  & .8869 & 82.81\% & 21.4 & 51.0  \\ \hline
LCN-gated & .8902 & 83.24\% & 22.3 & 55.8  \\

\end{tabular}
\caption{ Comparisons of various ways to create word lattice. l.qu and l.sen are the average token numbers in questions and sentences respectively. The 3 models in the middle construct lattices by adding words to CNN-char. +2\& considers the intersection of words of CTB and PKU mode while +2 considers the union. +20 uses the top 10 results of the two segmenters. 
\label{lattcie_construct}}
\end{table}

\subsubsection{Parameters and Efficiency}
LCNs only introduce inappreciable parameters in gated pooling besides the increasing vocabulary, which will not bring a heavy burden. The training speed is about 2.8 batches per second, 5 times slower than original CNNs, and the whole training of a 2-layer \textbf{LCN-gated} on DBQA dataset only takes about 37.5 minutes\footnote{Environment: CPU, 2*XEON E5-2640 v4. GPU: 1*NVIDIA GeForce 1080Ti}. The efficiency may be further improved if the network structure builds dynamically with supported frameworks. The fast speed and little parameter increment give LCNs a promising future in more NLP tasks.

\subsection{Case Study}

\begin{figure}[]
    \centering
    \includegraphics[width=0.47\textwidth]{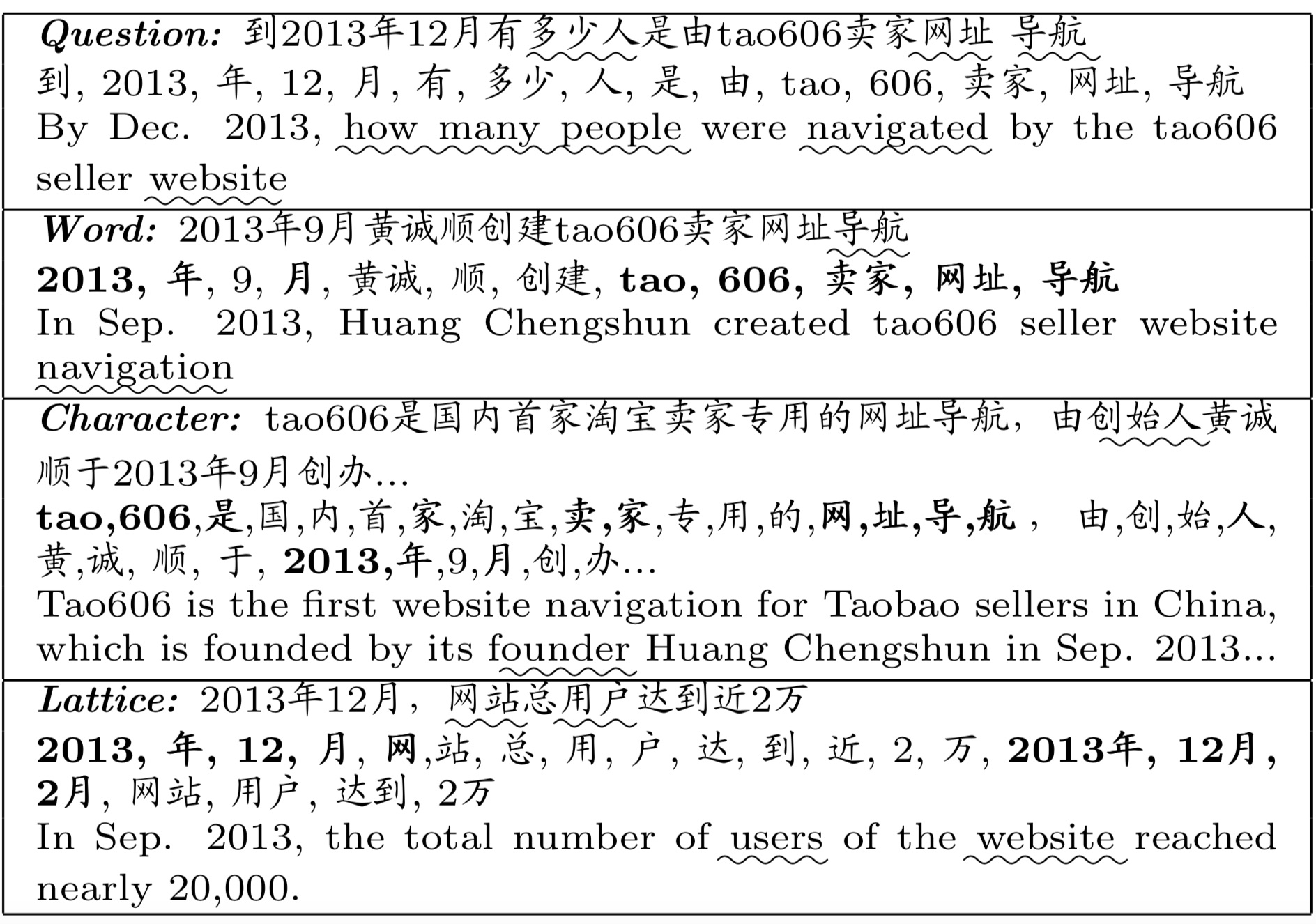}
    
\caption{ Example, a question (in word) and 3 sentences selected by 3 systems. {\bf Bold} means exactly sequence match between question and answer. Words with wave lines are mentioned in Section~Case Study.
        \label{case_study}}
\end{figure}

Figure~\ref{case_study} shows a case study comparing models in different input levels. The word level model is relatively coarse in utilizing informations, and finds a sentence with the longest overlap (5 words, 12 characters). However, it does not realize that the question is about numbers of people, and the \emph{``DaoHang''(navigate)} in question is a verb, but noun in the sentence. 
The character level model finds a long sentence which covers most of the characters in question, which shows the power of fine-granularity matching. But without the help of words, it is hard to distinguish the \emph{``Ren''}(people) in \emph{``DuoShaoRen''}(how many people) and \emph{``ChuangShiRen''}(founder), so it loses the most important information. 
While in lattice, although overlaps are limited, \emph{``WangZhan''}(website, \emph{``Wang''} web, \emph{``Zhan''} station) can match \emph{``WangZhi''}(Internet addresses, \emph{``Wang''} web, \emph{``Zhi''} addresses) and also relate to \emph{``DaoHang''}(navigate), from which it may infer that \emph{``WangZhan''}(website) refers to ``tao606 seller website navigation''(a website name).  Moreover, \emph{``YongHu''}(user) can match \emph{``Ren''}(people). With cooperations between characters and words, it catches the key points of the question and eliminates the other two candidates, as a result, it finds the correct answer.

\section{Related Work}
Deep learning models have been widely adopted in natural language sentence matching. 
Representation based models \cite{matchzooCDSSM,sa_qa1,sa_qa2,sa_qa3} encode and compare matching branches in hidden space.
Interaction based models \cite{matchzooMP,matchzooMVLSTM,sa_match1} incorporates interactions features between all word pairs and adopts 2D-convolution to extract matching features.
Our models are built upon the representation based architecture, which is better for short text matching.

In recent years, many researchers have become interested in utilizing all sorts of external or multi-granularity information in matching tasks.  
\cite{multigrancnn} exploit hidden units in different depths to realize interaction between substrings with different lengths. 
\cite{sa_match1} join multiple pooling methods in merging sentence level features, 
\cite{chen2018mix} exploit interactions between different lengths of text spans. 
For those more similar to our work, \cite{sa_match1} also incorporate characters, which is fed into LSTMs and concatenate the outcomes with word embeddings, and \cite{sa_qa3} utilize words together with predicate level tokens in KBRE task.
However, none of them exploit the multi-granularity information in word lattice in languages like Chinese that do not have space to segment words naturally. 
Furthermore, our model has no conflicts with most of them except \cite{sa_match1} and could gain further improvement.

GCNs\cite{gcn1,gcn3} and graph-RNNs\cite{peng2017cross,song2018graph} have extended CNNs and RNNs to model graph information,
and DGCs generalize GCNs on directed graphs in the fields of semantic-role labeling \cite{dgc1}, document dating \cite{dgc3}, and SQL query embedding\cite{xu2018sql}. However, DGCs control information flowing from neighbor vertexes via edge types, while we focus on capturing different contexts for each word in word lattice via convolutional kernels and poolings. 

Previous works involved Chinese lattice into RNNs for Chinese-English translation\cite{latticeLSTMnmt}, Chinese named entity recognition\cite{latticeLSTMner}, and Chinese word segmentation\cite{yang2018subword}.
To the best of our knowledge, we are the first to conduct CNNs on word lattice, and the first to involve word lattice in matching tasks.
And we motivate to utilize multi-granularity information in word lattices to relieve word mismatch and diverse expressions in Chinese question answering, while they mainly focus on error propagations from segmenters.

\section{Conclusions}

In this paper, we propose a novel neural network matching method (LCNs) for matching based question answering in Chinese. 
Rather than relying on a word sequence only, our model takes word lattice as input. 
By performing CNNs over multiple n-gram context to exploit multi-granularity information, LCNs can relieve the word mismatch challenges.
Thorough experiments show that our model can better explore the word lattice via convolutional operations and rich context-aware pooling, thus outperforms the state-of-the-art models and competitive baselines by a large margin. Further analyses exhibit that lattice input takes advantages of word and character level information, and the vocabulary based lattice constructor outperforms the strategies that combine characters and different word segmentations together.

\section{Acknowledgments}
This work is supported by Natural Science Foundation of China (Grant No. 61672057, 61672058, 61872294); the UK Engineering and Physical Sciences Research Council under grants EP/M01567X/1 (SANDeRs) and EP/M015793/1 (DIVIDEND); and the Royal Society International Collaboration Grant (IE161012). For any correspondence, please contact Yansong Feng.

% \section{Appendix}
% \subsection{Formula Derivation}
% \input{chap/appendix}

% References and End of Paper
% These lines must be placed at the end of your paper
% \bibliography{aaai}
% \bibliographystyle{aaai}

% \section{Appendix}
% \subsection{Formula Derivation}
% \input{chap/appendix}

%%%这后面是结尾
%%%这后面是结尾
%%%这后面是结尾

\end{document}